\documentclass[lettersize,journal]{IEEEtran}
\usepackage{amsmath,amsfonts}
\usepackage{algorithmic}
\usepackage{algorithm}
\usepackage{array}
\usepackage[caption=false,font=normalsize,labelfont=sf,textfont=sf]{subfig}
\usepackage{textcomp}
\usepackage{stfloats}
\usepackage{url}
\usepackage{verbatim}
\usepackage{graphicx}
\usepackage{cite}
\usepackage{multirow}
\usepackage{booktabs}
\usepackage[colorlinks,
            linkcolor=red,
            anchorcolor=black,
            citecolor=green
            ]{hyperref}
\begin{document}

\title{Occlusion-Aware 3D Hand-Object Pose Estimation\\ with Masked AutoEncoders}

\author{Hui Yang, Wei Sun, Jian Liu, Jin Zheng, Jian Xiao, and Ajmal Mian,~\IEEEmembership{Senior Member,~IEEE}
\thanks{This work is supported by the National Natural Science Foundation of China under Grants 62473141 and U22A2059, National Science and Technology Major Project of China under Grants 2026ZD1610900, Natural Science Foundation of Hunan Province under Grant 2024JJ5098, and the Open Foundation of the State Key Laboratory of Advanced Design and Manufacturing for Vehicle Body. Ajmal Mian was supported by the Australian Research Council Future Fellowship Award funded by the Australian Government under Project FT210100268. (\emph{Corresponding authors: Wei Sun; Jian Liu.})}
\thanks{Hui Yang, Wei Sun, Jian Liu, and Jian Xiao are with the National Engineering Research Center for Robot Visual Perception and Control Technology, School of Artificial Intelligence and Robotics, Hunan University, Changsha, 410082, China. (e-mail: \{huiyang, wei\_sun, jianliu, xiaojian2002\}@hnu.edu.cn)}
\thanks{Jin Zheng is with the School of Architecture and Art, Central South University, Changsha, 410082, China. (e-mail: zheng.jin@csu.edu.cn)}
\thanks{Ajmal Mian is with the Department of Computer Science and Software Engineering, The University of Western Australia, WA 6009, Australia. (e-mail: ajmal.mian@uwa.edu.au)}
}



\maketitle

\begin{abstract}
Hand-object pose estimation from monocular RGB images remains a significant challenge mainly due to the severe occlusions inherent in hand-object interactions.
Existing methods do not sufficiently explore global structural perception and reasoning, which limits their effectiveness in handling occluded hand-object interactions.
To address this challenge, we propose an occlusion-aware hand-object pose estimation method based on masked autoencoders, termed as HOMAE. Specifically, we propose a target-focused masking strategy that imposes structured occlusion on regions of hand-object interaction, encouraging the model to learn context-aware features and reason about the occluded structures. We further integrate multi-scale features extracted from the decoder to predict a signed distance field (SDF), capturing both global context and fine-grained geometry. To enhance geometric perception, we combine the implicit SDF with an explicit point cloud derived from the SDF, leveraging the complementary strengths of both representations. This fusion enables more robust handling of occluded regions by combining the global context from the SDF with the precise local geometry provided by the point cloud. Extensive experiments on challenging DexYCB and HO3Dv2 benchmarks demonstrate that HOMAE achieves state-of-the-art performance in hand-object pose estimation.
\end{abstract}


\begin{figure}
\centering
\includegraphics[width=1\linewidth]{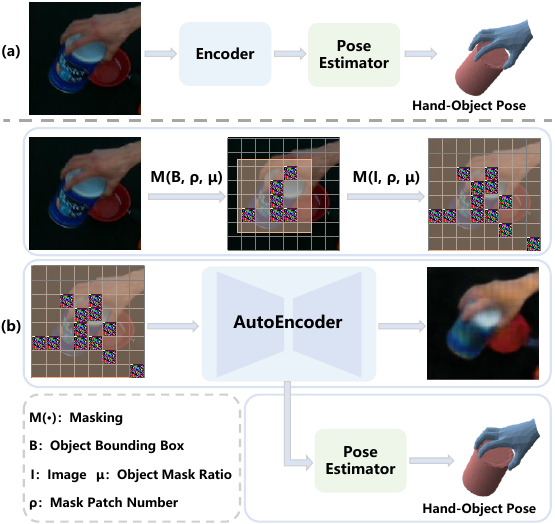}%
\caption{Comparison of existing methods with our HOMAE architecture. (a) Existing methods \cite{hasson2019learning}, \cite{tze}, \cite{qi2024hoisdf} extract features from the input image using an encoder and directly regress the hand-object pose through a pose estimator. (b) HOMAE introduces a target-focused masking mechanism during training by applying localized occlusions to the image and reconstructing the masked regions with an autoencoder. This design encourages the model to learn occlusion-aware representations, improving its understanding of occluded structures, and enhances the accuracy of hand-object pose estimation.}
\label{fig1}
\vspace{-1em}
\end{figure}

\section{Introduction}
\IEEEPARstart{H}{and}-object pose estimation from a single RGB image is a fundamental task in multimedia and computer vision, and plays a critical role in applications such as virtual reality \cite{chen2019overview}, \cite{keighrey2020physiology}, augmented reality \cite{lu2024spmhand}, robotics \cite{diff9d}, \cite{billard2019trends}, and human-robot interaction \cite{ren2020review}, \cite{liu2024survey}, \cite{song2022multimodal}.

\par Hand-object pose estimation has recently attracted considerable attention and achieved significant progress \cite{wang2023interacting}, \cite{chen2023tracking}, \cite{zhu2024contactart}, \cite{lj_tii}. However, robust estimation remains highly challenging, primarily due to severe self-occlusions of the hand and mutual occlusions during hand-object interactions. We argue that a key limitation of existing approaches
is due to their inability to model occlusions which hinders accurate perception and reasoning of occluded hand-object relationships, thereby limiting their ability to perform robust pose estimation.

\par Existing hand-object pose estimation methods can be broadly categorized into keypoint-based methods \cite{doosti2020hope},\cite{liu2021semi},\cite{hampali2022keypoint_transformer} and implicit 3D representation-based methods \cite{chen2022alignsdf},\cite{chen2023gsdf},\cite{qi2024hoisdf}.  
The former first detect 2D keypoints in RGB images and then utilize predefined 3D keypoints of objects, combined with the Perspective-n-Point (PnP) algorithm to estimate object poses. For hand pose estimation, keypoint-based methods predict 2D joint locations and shape parameters based on the MANO parameters model \cite{mano}. 
However, under severe occlusions during hand-object interactions, these methods often struggle to accurately locate the keypoints which result in incorrect feature associations, ultimately reducing the stability and accuracy of pose estimation. 
Implicit 3D representation-based methods utilize continuous functions such as the signed distance field (SDF) to encode the geometric structure of the hand and object. Although these methods provide improved geometric modeling capability, they heavily rely on visible regions. Under high occlusions, the lack of sufficient context leads to inaccurate SDF prediction, which hampers the perception of complete hand-object interaction structures, thereby limiting the accuracy of pose estimation.

\par \par To address the above challenges, we propose an occlusion-aware hand-object pose estimation framework that accurately reasons hand-object interaction relationship under severe occlusions, enabling precise pose estimation. A comparison of previous methods to our occlusion-awareness method is shown in Fig. \ref{fig1}. We propose an occlusion-aware hand-object pose estimation method based on masked autoencoders (MAE), termed HOMAE. First, we introduce a target-focused masking strategy that imposes structured occlusion on hand-object interaction regions, encouraging the masked autoencoder to learn occlusion-aware features and infer the occluded structures. 
Specifically, we identify hand-object interaction regions by leveraging object bounding boxes, and apply masking selectively within these regions to simulate realistic occlusions.
This compels the model to focus on the spatial relationships between the hand and object, thereby enhancing its ability to reason about interaction under occlusion.
To address the problem of inaccurate SDF  prediction under occlusion, we integrate hierarchical features from multiple decoding stages of the MAE to predict the SDF. This enables the model to capture both global structural consistency and fine-grained geometric details. Our design improves the accuracy of the SDF prediction by preserving high-frequency geometric information while incorporating essential contextual cues.
To further enhance geometric perception, we combine the implicit SDF with an explicit point cloud representation derived from the SDF, leveraging the complementary strengths of both representations. While the SDF encodes the global structure in a continuous manner, the explicit point cloud emphasizes structurally salient regions, providing precise local geometric cues. This fusion enables hand-object interaction modeling that is robust to occlusions by jointly capturing global context and local details.

\par Our contributions are summarized as follows:
\begin{itemize}
    \item We propose an occlusion-aware hand-object pose estimation method based on masked autoencoders to perceive and infer occluded regions in hand-object interactions by reconstructing masked input images, improving occlusion reasoning and interaction understanding ability.
    \item We introduce a multi-scale feature aggregation strategy, integrating hierarchical features from multiple decoding stages of the MAE to predict the SDF, capturing both global structure and fine-grained details for more accurate SDF prediction under occlusion.
    \item  To further improve occlusion-aware reasoning, we integrate both implicit and explicit representations by combining the predicted SDF with a point cloud derived from it. While the SDF encodes implicit global context, the derived point cloud provides explicit and localized geometric cues that are crucial for capturing fine-grained surface details. This complementary fusion enhances robust and accurate hand-object pose estimation.
\end{itemize}

\section{Related Work}

Since our method leverages the occlusion-awareness capabilities of MAE to address the occlusion challenge in 3D hand-object pose estimation, we categorize related work into two aspects: 3D hand-object pose estimation and masked autoencoders.

\subsection{3D Hand-Object Pose Estimation}
Previous researches have primarily focused on hand \cite{lu2024spmhand},  \cite{zhang2020differentiable}, \cite{zhou2023realistic} or object pose estimation \cite{mh6d}, \cite{rgb6d}, \cite{sinref6d}. With the development of large-scale hand-object interaction datasets\cite{hampali2020honnotate}, \cite{chao2021dexycb}, \cite{liu2022hoi4d}, \cite{zhu2024contactart}, increasing numbers of researchers have begun exploring hand-object interaction pose estimation \cite{hasson2019learning}, \cite{yang2022artiboost}, \cite{cai2025geometry}, \cite{xie2025maskhoi}. This has led to more accurate modeling of hand-object dynamics, driving advancements in robotic manipulation and human-robot interaction.
Existing hand-object pose estimation methods can be broadly categorized into two main approaches: keypoint-based methods \cite{doosti2020hope},\cite{liu2021semi},\cite{hampali2022keypoint_transformer},\cite{lin2023harmonious}, \cite{kuang2024learning} and implicit 3D representation-based methods \cite{chen2022alignsdf},\cite{chen2023gsdf},\cite{zhang2023ddf},\cite{qi2024hoisdf},\cite{jiang2024hand}.

\par For keypoint-based methods, Doosti \emph{et al.}\cite{doosti2020hope} proposed a lightweight deep learning framework to accurately predict hand-object poses from a single RGB image. Their framework employs a hand decoder to predict 2D joints and a 3D mesh parameterized by the MANO \cite{mano} model, while an object decoder estimates the 2D locations of predefined 3D corner points. The object pose is then recovered using the PnP algorithm.  
To address the challenge of obtaining ground-truth annotations in real-world scenarios, Liu \emph{et al.}\cite{liu2021semi} introduced a joint learning framework that leverages spatiotemporal consistency in large-scale hand-object videos as constraints for generating pseudo-labels in a semi-supervised learning paradigm. They further utilized a transformer-based \cite{vaswani2017attention} approach to perform explicit contextual reasoning between hand and object representations, thereby enhancing hand-object pose estimation.  
Hampali \emph{et al.}\cite{hampali2022keypoint_transformer} utilized a cross-attention mechanism to model the correlation between 2D keypoints and 3D hand-object poses.  
Lin \emph{et al.}\cite{lin2023harmonious} introduced a dual-stream backbone strategy that enables the hand and object to be extracted as distinct entities in intermediate layers, preventing feature competition during learning. The shared higher-level representations enforce feature harmonization between the hand and object, facilitating mutual feature enhancement.
To better exploit contextual information, Kuang \emph{et al.} \cite{kuang2024learning} proposed a context-aware transformer framework that jointly models global context and local hand–object features through shared decoder layers. By imposing mutual contextual priors between the hand and object, their method enhances robustness under occlusions.

\par For the implicit 3D representation-based methods, Chen \emph{et al.} \cite{chen2022alignsdf} proposed a joint learning framework for 3D hand-object reconstruction, integrating the advantages of parametric mesh models and SDF. Their approach estimates hand and object poses using a parametric model while leveraging an SDF network to learn hand and object shapes in a pose-normalized coordinate space. To better model the 3D geometry of hand-object interactions, Chen \emph{et al.}\cite{chen2023gsdf} further predicted kinematic chains for pose transformations and aligned SDF representations with highly articulated hand poses. By enforcing geometric alignment, their method improves the visual features of 3D points and enhances robustness against motion blur by incorporating temporal information. Qi \emph{et al.} \cite{qi2024hoisdf} introduced an SDF-guided hand-object pose estimation network that jointly utilizes hand and object SDFs to provide a global implicit representation over the complete reconstructed volume. Zhang \emph{et al.} \cite{zhang2023ddf} employed a deep distance field as an implicit shape representation. They proposed a 2D ray-based feature aggregation scheme and a 3D intersection-aware hand pose embedding to extract local features, effectively capturing hand-object interactions. Unlike above methods that solely rely on implicit 3D features, we further incorporate explicit 3D geometric features to enhance the representation of hand-object interactions.

\begin{figure*}
\centering
\includegraphics[width=1\linewidth]{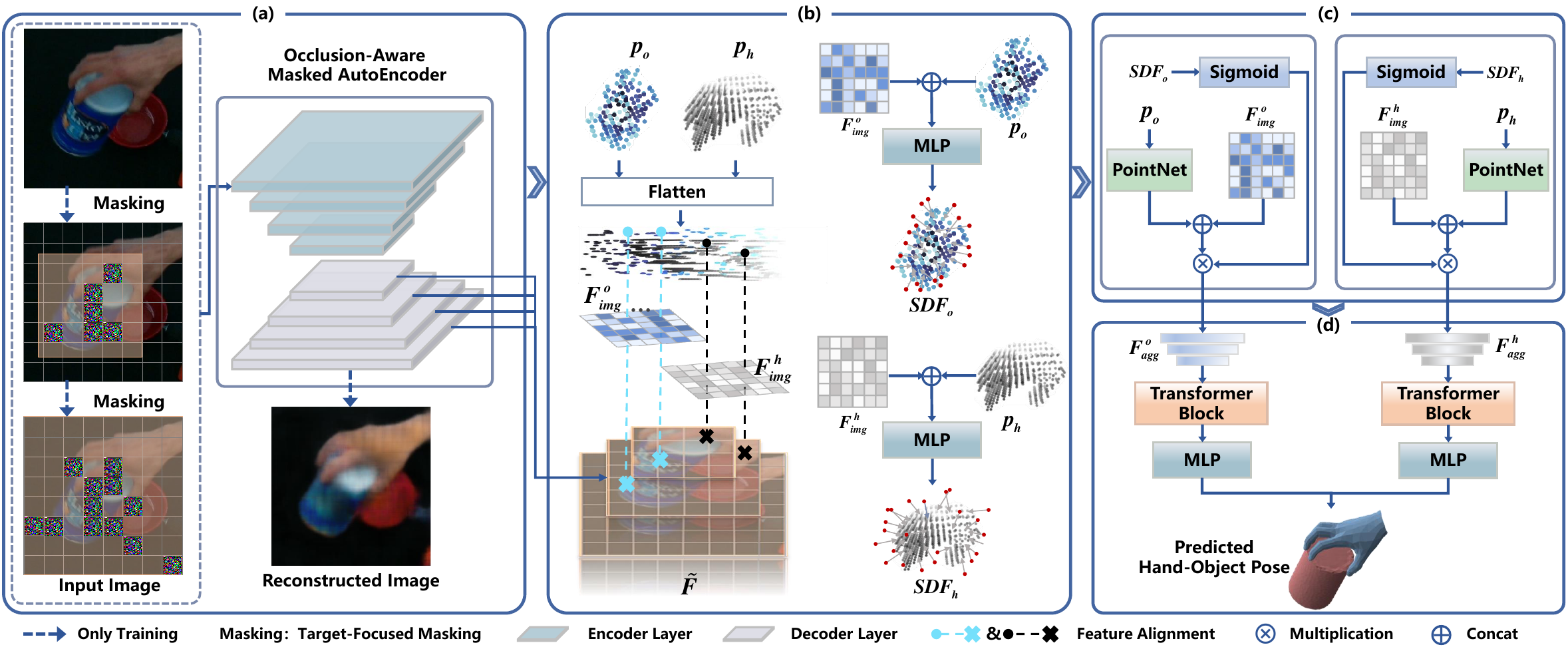}%
\vspace{-0.5em}
\caption{We propose HOMAE, a framework for estimating 3D hand-object pose from a single RGB image. The framework consists of four main components:
(a) MAE with target-focused masking: given an RGB image, a target-focused masking strategy is applied during training to guide the MAE in reconstructing the input image. During inference, no masking or reconstruction is required.
(b) Multi-scale feature extraction and SDF prediction: multi-scale image features are extracted from the decoder layers of the MAE and aligned point-wise with sampled hand-object point clouds. These aligned features are concatenated and passed through an MLP to predict the SDF. During training, the hand and object point clouds are sampled from mesh surfaces; during inference, the hand and object point clouds are voxel-sampled \cite{qi2024hoisdf} without requiring ground-truth meshes.
(c) Implicit–explicit geometric feature fusion: PointNet \cite{qi2017pointnet} is used to extract explicit geometric features from the hand and object point clouds. These are concatenated with aligned image features and element-wise multiplied with activated implicit SDF to generate fused implicit and explicit geometric representations.
(d) Hand-Object Pose estimation: the fused features of the hand and object are separately processed through transformer block followed by MLP to regress the final hand-object poses.}
\label{fig2}
\vspace{-1em}
\end{figure*}

\subsection{Masked Autoencoders}
MAE \cite{he2022mae} learn robust visual representations by randomly masking patches of the input image and training the model to reconstruct the masking patches. This mask-and-reconstruct mechanism enables the model to capture contextual dependencies and infer occluded structures, thereby enhancing its ability to reason under occlusions. Owing to its strong generalization and occlusion-aware capabilities, MAE has been successfully applied to a wide range of vision tasks.
In 2D vision tasks \cite{fan2024benchmarks} \cite{song2025diffcl}, \cite{qing2023mar}, Hu \emph{et al}. \cite{hu2024personmae} reconstructed the masked input image, forcing the model to capture all relevant features, thereby enhancing the reasoning ability of human images and better achieving the pedestrian re-identification task.
Bar \emph{et al.}\cite{bar2024egopet} leveraged MAE as a self-supervised learning paradigm to construct hand-object interaction datasets.
This global feature learning has shown to significantly enhance performance across various 2D vision downstream applications. MAE has also been increasingly adopted in 3D vision tasks \cite{qi2024shapellm}, \cite{xie2024template}, \cite{chen2023pimae}. By extending the masked autoencoding paradigm to 3D data, recent studies have explored various strategies for learning geometric representation. For instance, Mo \textit{et al.} \cite{mo2024fast} proposed a novel voxel-aware masking strategy that adaptively aggregates background/foreground information from voxelized point clouds, resulting in better point cloud generation. Xu \textit{et al.} introduce a masking mechanism over partial body joint coordinates and leverage spatiotemporal dependencies to recover the masking joints, thereby capturing richer relational cues for enhanced feature learning. These works demonstrate that MAE can effectively reconstruct 3D structures and learn geometry-aware representations, which are beneficial for downstream tasks in 3D vision.

\par We are the first to introduce MAE into hand-object pose estimation. This paper presents HOMAE, which explores the potential of MAE in this task and achieves state-of-the-art performance.

\section{Method}
We propose an occlusion-aware framework for hand-object pose estimation, as illustrated in Fig. \ref{fig2}, which aims to jointly estimate the hand-object pose from a single RGB image. Specifically, the regression targets include joint rotations $\theta \in \mathbb{R}^{3 \times 16}$ and a shape vector $\alpha \in \mathbb{R}^{10}$, as defined by the MANO model \cite{mano}, along with the 6-degree-of-freedom (6D) object pose, which includes a 3D rotation vector $r \in \mathbb{R}^3$ and a 3D translation vector $t \in \mathbb{R}^3$. Our approach consists of four key components: occlusion-aware masked autoencoders (Section \ref{section:III-A}), multi-scale feature-guided field regression (Section \ref{section:III-B}), implicit-explicit geometric aggregation (Section \ref{section:III-C}), and hand-object pose regression (Section \ref{section:III-D}).

\subsection{Occlusion-Aware Masked Autoencoders}
\label{section:III-A}

\par In monocular RGB-based hand-object interaction pose estimation, accurately extracting features of the hand and object remains challenging, especially under severe occlusions. To address this, we introduce an occlusion-aware learning mechanism to enhance feature reasoning capabilities in occlusion regions. Specifically, we propose a target-focused masking strategy that imposes structured occlusions on hand-object interaction regions. This design adaptively suppresses irrelevant background information while highlighting critical interaction regions, thereby guiding the model to focus on informative features during training and enhancing its ability to reason under occlusion. Furthermore, we incorporate a masked autoencoder for feature learning, where the encoder extracts semantically rich representations of hand-object interactions, and the decoder reconstructs occluded regions to reinforce the occlusion reasoning capabilities of the model. This joint encoding-decoding mechanism enables the model to focus on contextually relevant structures, thereby enhancing the quality of features in hand-object interaction and ultimately improving the accuracy of hand-object pose estimation.

\begin{figure*}
\centering
\includegraphics[width=0.85\linewidth]{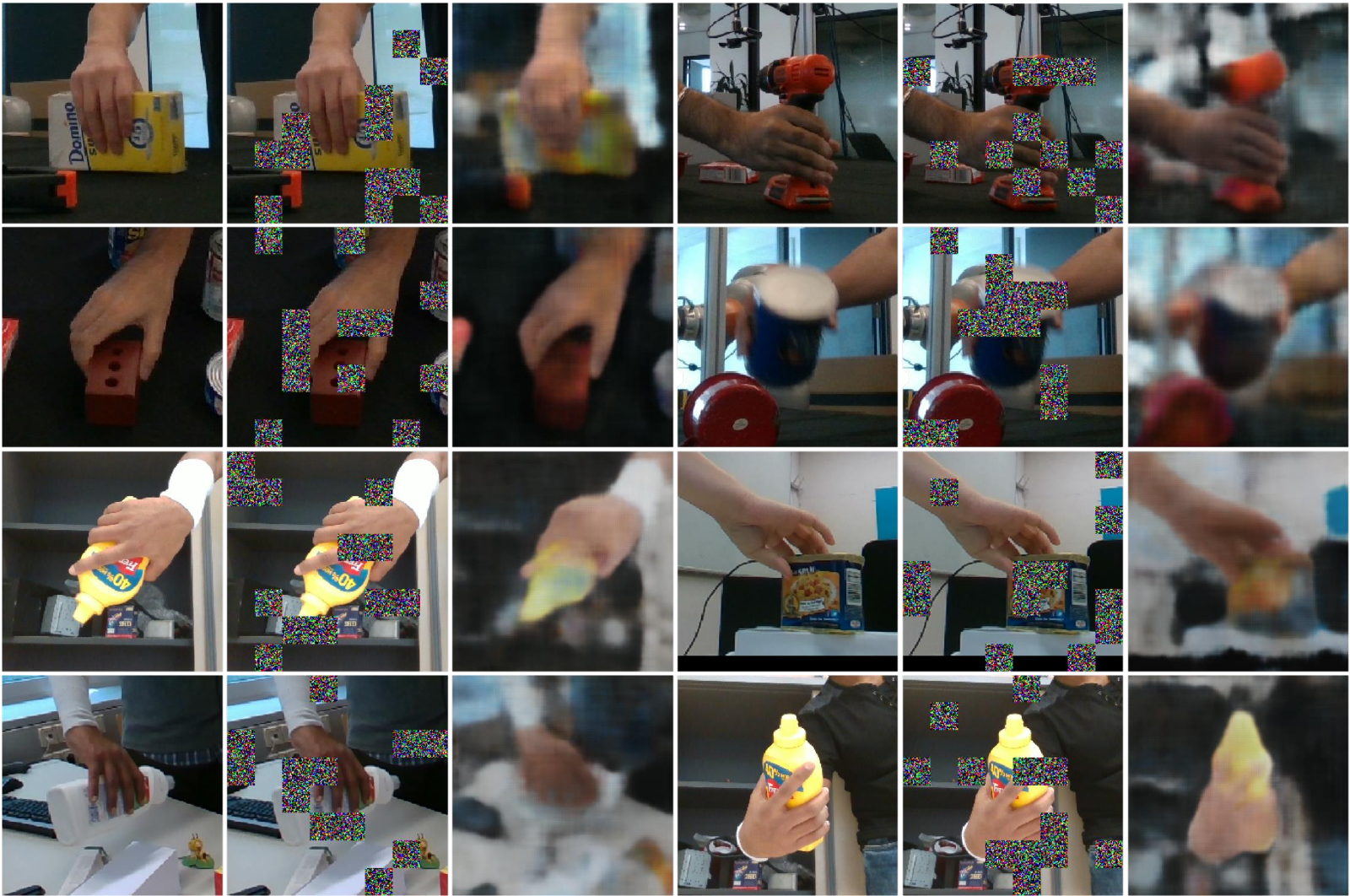}%
\vspace{-0.5em}
\caption{The reconstruction results of the masked images. Each group contains three images: the ground-truth image, the masked image, and our reconstructed image. The top two rows show results from the HO3Dv2 dataset \cite{hampali2020honnotate}, while the bottom two rows present results from the DexYCB dataset \cite{chao2021dexycb}.}
\label{fig3}
\vspace{-1em}
\end{figure*}

\par \textbf{Target-Focused Masking:} Given an input image \( I \in \mathbb{R}^{ H \times W \times 3} \), we divide it into patches of size \( P \times P \). We randomly mask the $\rho$ patches.
\begin{equation}
        B = (x_{\min}, y_{\min}, x_{\max}, y_{\max}),
    \end{equation}
 where \(B\) denotes the object bounding box. In the grid coordinate system, the patch range covering the object is:
\begin{equation}
    (x'_{\min}, y'_{\min}, x'_{\max}, y'_{\max}) = \frac{B}{P},
\end{equation}
   and the number of patches covering the object region is given by:
\begin{equation}
        N_o = (x'_{\max} - x'_{\min} + 1) \cdot (y'_{\max} - y'_{\min} + 1).
    \end{equation}

\par Randomly masking $\mu$ patches within the object region. The number of masked patches in the object region is given by:
 \begin{equation}
        N_m^o =  \mu \rho, 
    \end{equation}
the remaining patches are then randomly selected from the entire image, excluding those already chosen in the object region: 
\begin{equation}
        N_m^b = \rho  - N_m^o,
\end{equation}
    where the \( N_m^b \) patches are randomly sampled from the entire image, including both the object region and the background. The final mask matrix \( M \) has the following shape:
\begin{equation}
    M_{i,j} =
    \begin{cases}
        1, & \text{if masked} \\
        0, & \text{otherwise}
    \end{cases}, \quad M \in \{0,1\}^{\frac{H}{P} \times \frac{W}{P}}.
\end{equation}

\par After determining the mask matrix $M$, we generate the masked image \( I_m \in \mathbb{R}^{ H \times W \times 3} \) by replacing each masked patch $(i, j)$ in $I$ with a Gaussian noise block sampled randomly: 
\begin{equation}
    I_m[:, i P: (i+1) P, j P: (j+1) P] = \mathcal{N}(0,1),
\end{equation}
where \( \mathcal{N}(0,1) \) represents Gaussian noise sampled independently for each masked patch.

\par \textbf{Autoencoder:}  
After obtaining the masking image \( I_m \), we employ DINOv2 \cite{DINOv2} as the encoder \( \mathcal{E} \) to fully leverage its powerful feature extraction capability for capturing global semantic information. Pre-trained through self-supervised learning, DINOv2 effectively extracts structured features from images, enabling it to focus on key regions even in complex backgrounds and occluded scenarios. Formally, the extracted feature representation is given by:  
\begin{equation}
F = \mathcal{E}(I_m),
\end{equation}
where $F \in \mathbb{R}^{\frac{H}{14} \times \frac{W}{14} \times 256}$ represents the high-level semantic features extracted by the encoder $\mathcal{E}$. To enhance the model robustness to occlusion and preserve fine-grained interaction cues, we employ a decoder $\mathcal{D}$ composed of a stack of MLP layers, which progressively upsamples $F$ through $L$ decoding layers to generate multi-scale features for reconstructing the masked image:
\begin{equation}
    \hat{I} = \mathcal{D}(F) = \prod_{l=1}^{L} \mathcal{D}_l (F),
\end{equation}
where $\prod$ denotes the sequential decoding operations that generate multi-scale feature maps and reconstruct the complete image.

\par To enforce accurate reconstruction, we employ the mean squared error loss to supervise the predicted image \( \hat{I} \) against the ground-truth image \( I \):  
 \begin{equation}  
 L_{\text{rec}} = \frac{1}{\mathbb{P}} \sum_{i=1}^{\mathbb{P}} \| \hat{I}_i - I_i \|^2,  
 \end{equation}  
where \( \mathbb{P} \) denotes the total number of pixels. This loss ensures that the reconstructed image preserves structural consistency with the ground truth. This encoder-decoder framework not only enhances the understanding of hand-object interaction regions but also significantly improves occlusion awareness.
The visualization results of the masking and reconstructed images are shown in Fig.~\ref{fig3}. Our target-focused masking strategy effectively covers a wide range of occlusion types commonly encountered in real-world hand-object interactions, encompassing both fine-grained occlusions (e.g., fingertips) and large-area occlusions (e.g., palms or objects). Meanwhile, our method demonstrates strong reconstruction robustness under severe occlusions, effectively preserving the spatial relationships between the hand and object while recovering interaction regions with clear structural consistency.

\subsection{Multi-Scale Feature-Guided Field Regression}
\label{section:III-B}

To address the challenge of inaccurate SDF prediction under occlusions, we leverage multi-scale image features to enhance the implicit representation of hand-object interactions, thereby enabling more accurate and robust SDF estimation. The SDF encodes the distance of each point to the hand and object nearest surface, providing a continuous and differentiable representation of object and hand geometry. By integrating hierarchical image features, our method improves the robustness of SDF estimation, effectively capturing local texture details and global semantic cues.

\par \textbf{Feature Alignment:}  
To enhance SDF estimation, we exploit hierarchical features by aggregating intermediate outputs from multiple stages of the decoder. This allows the model to capture both fine-grained local textures and high-level semantic information. Formally, given the encoder output feature $F$, the decoder produces a multi-scale feature representation $\tilde{F} \in \mathbb{R}^{H \times W \times C}$, which integrates information from different decoding levels. 
\begin{equation}
    \tilde{F} = \text{MLP} \left( \bigoplus_{l \in L} \mathcal{D}^{(l)}(F) \right),
\end{equation}
where $\bigoplus$ denotes the channel-wise concatenation of features from different decoding stages.
This hierarchical feature representation captures comprehensive multi-scale information beneficial for predicting SDF.

\par Given the 3D surface points $p$ of the hand and object, we project them onto the 2D image plane using the camera intrinsic matrix \( K \). During training, the surface points are directly sampled from the ground-truth hand and object mesh. For inference, we sample potential surface points within the voxel space following HOISDF \cite{qi2024hoisdf} without requiring ground-truth meshes. To bridge the distribution gap between training and inference, clean mesh-sampled surface points are used for SDF prediction to ensure accurate signed distance learning, while during SDF feature learning, Gaussian-perturbed surface points are fed into the network to simulate the voxel-sampled distribution encountered at inference, thereby improving robustness.
\begin{equation}
    F_{\text{img}}^{x} = \tilde{F} \left( \pi(K, p_{x}) \right), \quad x \in \{h, o\},
\end{equation}
where \(\pi(K, p)\) represents the projection of the 3D point \(p\) onto the 2D image plane.
This process ensures that the 3D geometric structure is effectively aligned with image semantic information, facilitating more accurate SDF estimation in the subsequent regression module.

\par \textbf{SDF Regression:} For hand SDF estimation, in order to preserve the geometric properties of the original 3D structure, we apply Fourier Positional Encoding \cite{FLY} to the hand surface points  \( p_h \in \mathbb{R}^{600 \times 3} \), obtaining a high-dimensional representation \( \gamma(p_h) \in \mathbb{R}^{600 \times 30} \). The final feature representation for each surface point \( p_h \) is constructed by fusing the fourier encoded, pixel-aligned image features $F^h_{\text{img}} \in \mathbb{R}^{600 \times C} $, and the original 3D 
hand points. This fused representation is then fed into an MLP to regress the SDF for the hand surface:
\begin{equation}
   SDF_h = \text{MLP} \left( \gamma(p_h) \oplus F^h_{\text{img}}\oplus p_h \right). 
\end{equation}
The process for estimating the $SDF_o$ follows the same steps as the $SDF_h$, with \( p_o \in \mathbb{R}^{200 \times 3} \), $\gamma(p_o) \in \mathbb{R}^{200 \times 30} $, and \( F^o_{\text{img}} \in \mathbb{R}^{200 \times C}\) represent the object surface points and features. By leveraging hierarchical multi-scale image features and geometric priors, our method enables accurate SDF estimation for both the hand and object surfaces, even under occlusions and complex hand-object interactions.

\subsection{Implicit and Explicit Geometric Aggregation}
\label{section:III-C}
To achieve robust hand-object pose estimation under occlusion, we integrate implicit and explicit geometric representations. Implicit representations provide continuous, differentiable surface encoding, while explicit 3D point clouds offer structured spatial cues. Combined with multi-scale image features from the masked autoencoder, our model learns both local geometric structures and rich contextual information, enhancing its ability to infer accurate, occlusion-aware hand-object poses.

\par To extract point-wise geometric features, the sampled hand point clouds is fed into a PointNet \cite{qi2017pointnet} backbone, producing global feature \( F_{\text{3D}}^{\text{h}} \in \mathbb{R} ^{600 \times C} \) that encodes the structural characteristics of each point. Simultaneously, multi-scale image features \( F^h_{\text{img}} \) corresponding to the 2D projection of \( p_h \) are obtained from the decoder layers of the masked autoencoder. The final aggregated feature $F_{\text{agg}}^{\text{h}} \in \mathbb{R}^{600 \times C}$ for each hand point is obtained by concatenating its corresponding image feature and explicit 3D geometric feature, and then further fused with the implicit representation derived from the predicted SDF, enabling a richer encoding of both visual, geometric, and implicit spatial information.
\begin{equation}
   F_{\text{agg}}^{\text{h}} = \left( F^h_{\text{img}} \oplus F_{\text{3D}}^{\text{h}} \right) \cdot \frac{1}{\beta_h} \cdot \sigma_h\left(\frac{SDF_{\text{h}}(p_h)}{\beta_h}\right), 
\end{equation}
Where \( \sigma_h(\cdot) \) is the sigmoid function, \( \beta_h\) is the learnable scale parameter.
The object aggregated feature \( F_{\text{agg}}^{\text{o}} \in \mathbb{R}^{200 \times C} \) is obtained using the same strategy as for the hand.
By unifying implicit SDF encoding, explicit point cloud representations, and dense image features within a multi-modal fusion framework, our approach leverages the implicit geometric cues learned from the SDF to adaptively modulate and enhance explicit spatial features. This modulation mechanism strengthens structural reasoning under occlusion, leading to more robust and accurate hand-object pose estimation.

\subsection{Hand-Object Pose Estimation}
\label{section:III-D}
To estimate the hand and object poses, we utilize their respective aggregated features, \( F_{\text{agg}}^{\text{h}} \) and \( F_{\text{agg}}^{\text{o}} \), which fuse implicit and explicit geometric information with multi-scale semantics cues. These representations capture rich spatial context, enabling robust pose estimation under occlusions and complex hand-object interactions.
For hand pose estimation, \( F_{\text{agg}}^{\text{h}} \) is fed into a transformer block \cite{vaswani2017attention}. The refined features are then passed through an MLP to predict the MANO parameters \cite{mano}, including joint rotations and shape vector:
\begin{equation}
 (\{\theta_i \}_{i=0}^{16}, \alpha) = \text{MLP}_{\text{h}}(\text{Transformer}_{\text{h}}(F_{\text{agg}}^{\text{h}})),   
\end{equation}
where \( \theta\in \mathbb{R}^{3 \times 16} \) denote joint rotations and \( \alpha \in \mathbb{R}^{10} \) is the shape vector. 
For object pose estimation, the aggregated object feature \( F_{\text{agg}}^{\text{o}} \) is processed by a transformer block, followed by an MLP that outputs the object 6D pose parameters:
\begin{equation}
  (r, t) = \text{MLP}_{\text{o}}(\text{Transformer}_{\text{o}}(F_{\text{agg}}^{\text{o}})),  
\end{equation}
where \( r \in \mathbb{R}^3 \) and \( t \in \mathbb{R}^3 \) represent rotation and translation vector, respectively.

\textbf{Loss Function: } 
Our overall training objective combines multiple loss functions to jointly optimize image reconstruction quality and pose estimation accuracy. Specifically, a reconstruction loss $L_{\text{rec}}$ is employed to supervise masked image reconstruction. The hand pose loss $L_{\text{mano}}$ and the object pose loss $L_{\text{obj}}$, both implemented as smooth L1 losses, are used to constrain the predicted MANO parameters and object pose, respectively. Additionally, the SDF loss $L_{\text{SDF}}$ supervises the predicted implicit surface representation. The auxiliary loss terms $L_{\text{others}}$ follow HOISDF \cite{qi2024hoisdf}, providing additional regularization or task-specific constraints. The total loss is defined as: 
\begin{equation}
L_{\text{total}} = \lambda_{\text{1}} L_{\text{rec}} + \lambda_{\text{2}} L_{\text{mano}} + \lambda_{\text{3}} L_{\text{obj}} + \lambda_{\text{4}} L_{\text{SDF}} + \lambda_{\text{5}} L_{\text{others}}.
\end{equation}
\par Each term is weighted by its corresponding coefficient \( \lambda \), to balance the contributions to the overall loss.

\begin{table}[]
\renewcommand\arraystretch{1.25}
  \centering
\caption{Comparison with state-of-the-art hand-object pose estimation methods on the DexYCB dataset \cite{chao2021dexycb}. The best results are highlighted in bold.}
\begin{tabular}{c|cc|ccc}
\toprule[1.5pt]
Metrics in {[}mm{]} & MJE  & PAMJE & OCE  & MCE  & ADD-S  \\ \hline
Hasson \emph{et al.} \cite{hasson2019learning}      & 17.6 & -     & -    & -    & -       \\
Hasson \emph{et al.} \cite{hasson2021towards}      & 18.8 & -     & -    & 52.5 & -        \\
Tze \emph{et al.} \cite{tze}          & 15.3 & -     & -    & -    & -      \\
Li \emph{et al.} \cite{yang2022artiboost}           & 12.8 & -     & -    & -    & -         \\
Chen \emph{et al.} \cite{chen2022alignsdf}         & 19.0 & -     & 27.0 & -    & -         \\
Chen \emph{et al.} \cite{chen2023gsdf}         & 14.4 & -     & 19.1 & -    & -        \\
Wang \emph{et al.} \cite{wang2023interacting}        & 12.7 & 6.86  & 27.3 & 32.6 & 15.9      \\
Lin \emph{et al.} \cite{lin2023harmonious}         & 11.9 & 5.81  & 39.8 & 45.7 & 31.9    \\
Qi \emph{et al.} \cite{qi2024hoisdf}          & \textbf{10.1} & 5.31  & 18.4 & 27.4 & 13.3     \\ 
Cai \emph{et al.} \cite{cai2025geometry}          & 13.7 &  5.65 &-  & - & -    \\ 
Kuang \emph{et al.} \cite{kuang2024learning}          &11.1  & 5.30  &-  & - &  -    \\ \hline
Ours                &  10.6    & \textbf{5.08}      &  17.1   &\textbf{25.2}     &\textbf{ 11.8 }       \\
\bottomrule[1.5pt]
\end{tabular}
\label{dexycb}
\end{table}

\begin{figure}
\centering
\includegraphics[width=1\linewidth]{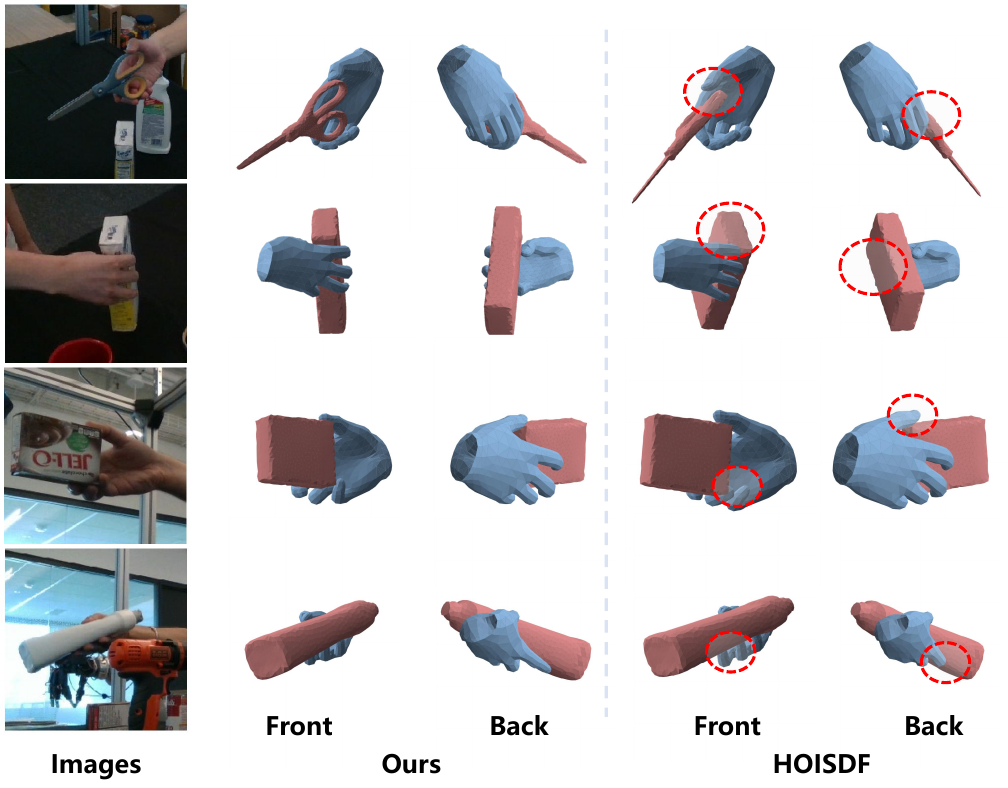}%
\caption{Qualitative comparison of hand-object pose estimation results by  HOMAE and HOISDF \cite{qi2024hoisdf} on the DexYCB dataset \cite{chao2021dexycb}. Front and back denote the front view and rear view, respectively. The red dotted circle indicates that HOISDF has lower pose estimation accuracy than our method.}
\label{fig4}
\vspace{-0.5em}
\end{figure}

\begin{table*}[t]
\renewcommand\arraystretch{1.25}
   \setlength{\tabcolsep}{4.1pt}
\centering
\caption{Quantitative comparison with hand mesh metrics on the DexYCB dataset \cite{chao2021dexycb}. $\uparrow$ indicates that higher values represent better performance, while $\downarrow$ means lower values are better.}
\begin{tabular}{c|ccccc|ccccc}
\toprule[1.5pt]
Methods                                              & PA-J-AUC$\uparrow$ & PA-V-PE$\downarrow$ & PA-V-AUC$\uparrow$ & PA-F@5$\uparrow$ & PA-F@15$\uparrow$ & J-AUC$\uparrow$ & V-PE$\downarrow$ & V-AUC$\uparrow$ & F@5$\uparrow$ & F@15$\uparrow$ \\ \hline
HandOccNet\cite{park2022handoccnet} & 88.4               & 5.5                 & 89.0               & 78.0             & \textbf{99.0}     & 74.8            & 13.1             & 76.6            & 51.5          & 92.4           \\
MobRecon\cite{chen2022mobrecon}     & 87.3               & 5.6                 & 88.9               & 78.5             & 98.8              & 73.7            & 13.1             & 76.1            & 50.8          & 92.1           \\
H2ONet \cite{akiva2021h2o}          & 88.9               & 5.5                 & 89.1               & 80.1             & \textbf{99.0}     & 74.6            & 13.0             & 76.2            & 51.3          & 92.1           \\
UniHOPE \cite{wang2025unihope}      & -                  & 5.4                 & -                  & -                & -                 & -               & 12.2             & -               & -             & -              \\ \hline
Ours                                                 & \textbf{89.8}      & \textbf{4.9}        & \textbf{90.0}      & \textbf{81.7}    & 98.9              & \textbf{79.7}   & \textbf{10.5}    & \textbf{79.9}   & \textbf{59.7} & \textbf{94.4}  \\ 
\bottomrule[1.5pt]
\end{tabular}
\label{hand_recon}
\end{table*}

\begin{table}[]
\renewcommand\arraystretch{1.25}
\centering
\caption{Comparison with state-of-the-art hand-object pose estimation methods on the HO3Dv2 dataset \cite{hampali2020honnotate}.}
\begin{tabular}{c|ccc|cc}
\toprule[1.5pt]

Metrics in {[}mm{]} & MJE  & STMJE & PAMJE & OME  & ADD-S \\ \hline
Hasson \emph{et al.} \cite{hasson2019learning}       & -    & 31.8  & 11.0  & -    & -     \\
Hasson \emph{et al.} \cite{hasson20_handobjectconsist}       & -    & 36.9  & 11.4  & 67.0 & 22.0  \\
Hasson \emph{et al.} \cite{hasson2021towards}      & -    & 26.8  & 12.0  & 80.0 & 40.0  \\
Liu \emph{et al.} \cite{liu2021semi}          & -    & 31.7  & 10.1  & -    & -     \\
Hampali \emph{et al.} \cite{hampali2022keypoint_transformer}      & 25.5 & 25.7  & 10.8  & 68.0 & 21.4  \\
Lin \emph{et al.} \cite{lin2023harmonious}         & 28.9 & 28.4  & \textbf{8.9}   & 64.3 & 32.4  \\

Qi \emph{et al.} \cite{qi2024hoisdf}          & 23.6 & 22.8  & 9.6   & 48.5 & 17.8  \\ 
Cai \emph{et al.} \cite{cai2025geometry}          & - & -  & 9.1 &-  & -    \\ 
Kuang \emph{et al.} \cite{kuang2024learning}          & - & 23.7  & \textbf{8.9} & - &  -    \\ \hline
Ours                &   \textbf{21.8}  & \textbf{20.5}    &   9.8    &  \textbf{39.3}    &\textbf{14.2}     \\ 
\bottomrule[1.5pt]
\end{tabular}
\label{ho3d}
\end{table}

\begin{table}
\renewcommand\arraystretch{1.25}
\centering
\caption{Per-object performance on HO3Dv2 dataset \cite{hampali2020honnotate}. Our method can outperform HOISDF \cite{qi2024hoisdf} on HO3Dv2 dataset as well.}
\begin{tabular}{ccccc}
\toprule[1.5pt]
\multirow{2}{*}{\begin{tabular}[c]{@{}c@{}}Method\\ Metrics in {[}mm{]}\end{tabular}} & \multicolumn{2}{c}{Ours}      & \multicolumn{2}{c}{Qi \cite{qi2024hoisdf}} \\ \cline{2-5} 
 & OME           & ADD-S         & OME         & ADD-S        \\ \hline
006 mustard bottle                                                                    & \textbf{42.3} & \textbf{11.2} & 52.1        & 13.1         \\
010 potted meat can                                                                   & \textbf{41.8} & \textbf{13.9} & 51.8        & 18.2         \\
021 bleach cleanser                                                                   & \textbf{33.8} & \textbf{17.6} & 41.6        & 22.2         \\ \hline
Mean                                                                                  & \textbf{39.3} & \textbf{14.2} & 48.5        & 17.8         \\ 
\bottomrule[1.5pt]
\end{tabular}
\label{tab3}
\end{table}

\section{Experiments}
To demonstrate the effectiveness of our approach, we compare with state-of-the-art methods on two benchmark datasets with occlusion challenges, DexYCB \cite{chao2021dexycb} and HO3Dv2 \cite{hampali2020honnotate}. Furthermore, we conduct ablation studies to evaluate the advantages of each proposed innovation.

\subsection{Datasets and Evaluation Metrics}
\label{sec:datasets}
\par \textbf{DexYCB dataset} is a large-scale benchmark for hand-object interaction pose estimation. The dataset captures 10 subjects manipulating 20 YCB objects \cite{xiang2017posecnn}. Following the standard evaluation protocol, we adopt the S0 split, using sequences from 8 subjects for training and the remaining 2 subjects for testing to ensure fair comparison with existing methods. To comprehensively evaluate both hand and object pose estimation performance, we adopt Mean Joint Error (MJE) and Procrustes-Aligned MJE (PA-MJE) \cite{PAMJE} for assessing hand pose accuracy, and Object Center Error (OCE), Mean Corner Error (MCE), and the average closest point distance (ADD-S) for evaluating object pose accuracy. Additionally, we report Mean Mesh Error (MME), the area under the curve of the percentage of correct vertices (V-AUC), and F-scores at 5mm and 15 mm thresholds (F@5mm, F@15mm), together with their Procrustes-aligned versions following H2ONet \cite{xu2023h2onet} to assess the quality of hand mesh reconstruction.

\par \textbf{HO3Dv2 dataset} comprises 77K images from 68 video sequences, covering 10 different subjects interacting with 10 objects from the YCB dataset \cite{xiang2017posecnn}. We follow the official data split protocol for training and testing and submit our test results to the official evaluation server. Performance is assessed using five key metrics: Mean Joint Error (MJE), Scale-Translation aligned MJE (STMJE) \cite{stnje}, and Procrustes-Aligned MJE (PA-MJE) \cite{PAMJE} for evaluating hand pose estimation, while Object Mesh Error (OME) and ADD-S are used to assess object pose estimation accuracy.

\begin{figure}
\centering
\vspace{-0.5em}
\includegraphics[width=1\linewidth]{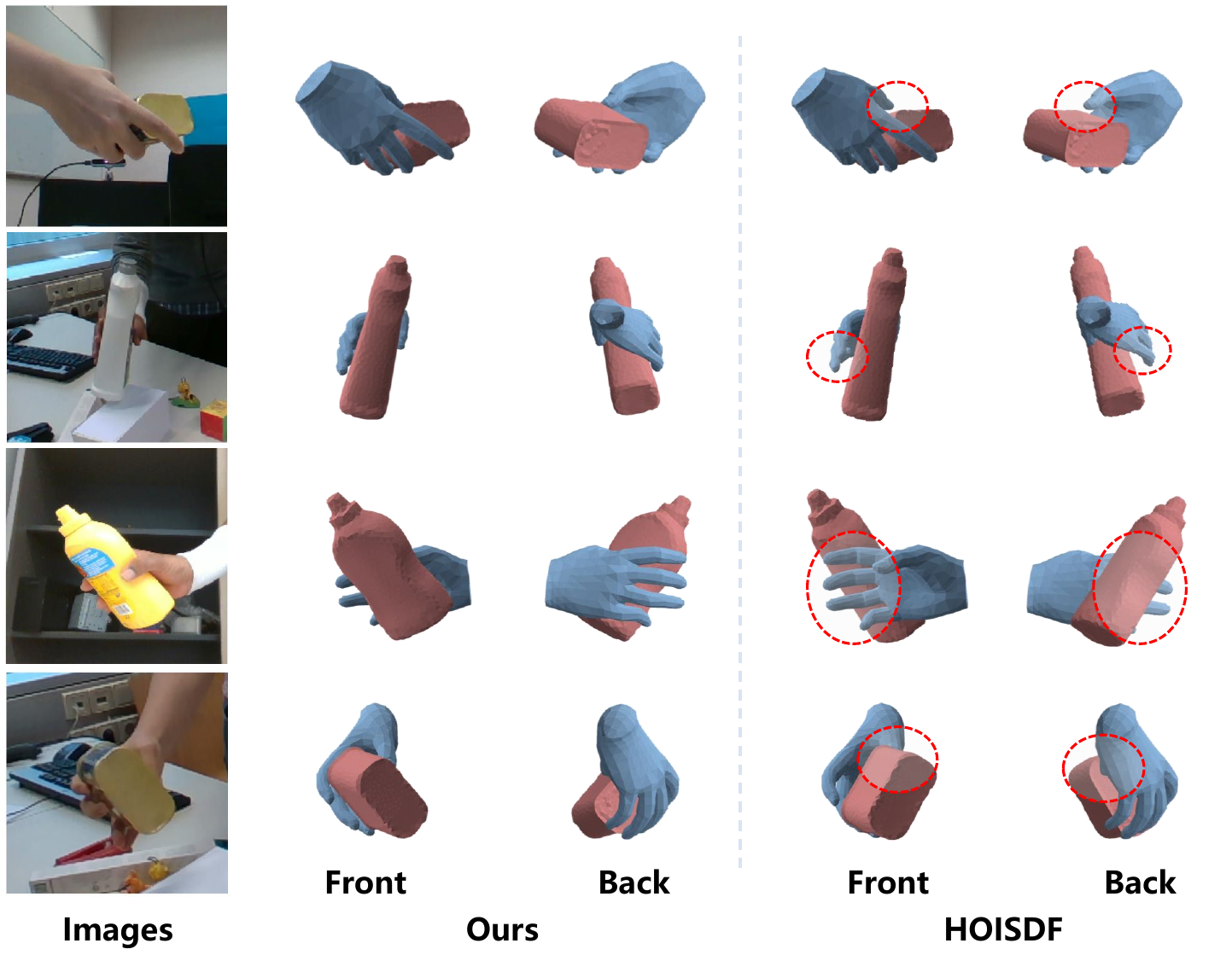}%
\caption{Qualitative comparison of hand-object pose estimation results by HOMAE and HOISDF \cite{qi2024hoisdf} on the HO3Dv2 dataset \cite{hampali2020honnotate}. The red dotted circle indicates that HOISDF has lower pose estimation accuracy than our method.}
\label{fig5}
\vspace{-1em}
\end{figure}

\subsection{Implementation Details}
\label{sec:implementation}
HOMAE is implemented in PyTorch and trained and inferred on a single NVIDIA RTX 3090 GPU. We adopt the Adam optimizer with a batch size of 24. The initial learning rate is set to 1e-4 and decayed by a factor of 0.7 every 5 epochs. The model is trained for 60 epochs on both the DexYCB \cite{chao2021dexycb} and HO3Dv2 \cite{hampali2020honnotate} datasets, which is sufficient to achieve satisfactory performance. The input images are cropped around the object and resized to 224×224. In our masking strategy, the patch size $P$ is set to 28, the total number of masks $\rho$ is 12, and the probability $\mu$ of a mask falling within the object bounding box is set to 50\%.  The feature dimension C is set to 223. Through extensive experiments, we found that setting $\lambda_1$, $\lambda_2$, $\lambda_3$, and $\lambda_4$ at 1 yields the best performance, while $\lambda_5$ is set according to the configuration in HOISDF \cite{qi2024hoisdf}.

\begin{table*}[]
\renewcommand\arraystretch{1.25}
\centering
\caption{Ablation studies on key components of our method on the HO3Dv2 dataset \cite{hampali2020honnotate}.}
\begin{tabular}{l|ccc|cc}
\toprule[1.5pt]
\multicolumn{1}{c|}{Metric in {[}mm{]}} & MJE & STMJE & PAMJE & OME &ADD-S \\ \hline
w/o Target-Focused Masking and Image Reconstruction                   &23.2      & 22.6     &10.5      &43.8     &15.9    \\
w/o Multi-Scale Feature Fusion for SDF Regression                 &22.4     & 21.5   &10.1     &41.2     & 14.9   \\
w/o Implicit and Explicit Geometric Aggregation (only implicit)           &22.1   &21.3     &10.3     & 41.6     &15.4  \\
w/o Noise Perturbation on Mesh-Sampled Surface Points       & 24.5 & 22.7 & 10.6 & 43.5 & 16.2 \\
DINOv2 $\rightarrow$ DINOv3 (Backbone Substitution)      & 22.3 & 20.9 & 10.2 & 39.8 & 14.5 \\ \hline
Ours (Full)   & \textbf{21.8}&\textbf{20.5} & \textbf{9.8}& \textbf{39.3}& \textbf{14.2}\\ \bottomrule[1.5pt]
\end{tabular}
\vspace{-1.5em}
\label{table:main_part}
\end{table*}

\begin{table}
\renewcommand\arraystretch{1.25}
\centering
\caption{Ablation study on loss functions on the HO3Dv2 dataset \cite{hampali2020honnotate}.}
\begin{tabular}{l|ccc|cc}
\toprule[1.5pt]
\multicolumn{1}{c|}{Metric in {[}mm{]}} & MJE & STMJE & PAMJE & OME & ADD-S \\ \hline
w/o $L_{\text{rec}}$  & 22.7 & 21.4 & 10.4 & 41.8 & 15.1 \\ 
w/o $L_{\text{SDF}}$  &128.6  & 112.9 & 29.6 & 95.3 & 87.6 \\ 
w/o $L_{\text{mano}}$     & 135.8 & 127.2 & 34.6 & 40.3 & 14.8 \\ 
w/o $L_{\text{obj}}$         & 22.1 & 20.6 & 10.1 & 262.2 & 273
.5\\ \hline
Ours (Full)                      & \textbf{21.8} & \textbf{20.5} & \textbf{9.8} & \textbf{39.3} & \textbf{14.2} \\ 
\bottomrule[1.5pt]
\end{tabular}
\vspace{-0.5em}
\label{table:main_part2}
\end{table}

\subsection{Comparison with State-of-the-Art Methods}
\label{sec:sota_comparison}
\par \textbf{Quantitative comparisons on DexYCB Dataset: }Table~\ref{dexycb} presents a comparison of our method with other approaches on the DexYCB dataset. The results demonstrate that our method achieves state-of-the-art performance across nearly all evaluation metrics. While achieving a competitive MJE of 10.6 mm, slightly behind Qi {\emph{et al.} \cite{qi2024hoisdf}(10.1 mm), our method achieves the best PAMJE of 5.08 mm, indicating more accurate joint localization after alignment. More importantly, our method significantly outperforms all previous methods on object mesh accuracy, obtaining the lowest OCE of 17.1 mm and MCE of 25.2 mm. Furthermore, we achieve the highest accuracy object pose estimation, with an ADD-S of 11.8 mm. These results highlight the strength of our approach in modeling fine-grained hand-object interactions, ensuring precise articulation and detailed geometric reconstruction, especially in challenging contact and occlusion scenarios. The qualitative comparison between our method and HOISDF \cite{qi2024hoisdf} on the DexYCB dataset is shown in Fig. \ref{fig4}. The visualization results demonstrate that our method achieves more accurate pose estimation under occluded scenarios, leading to more realistic and plausible hand-object contact. In contrast, HOISDF often fails to handle occlusions effectively, resulting in incorrect hand-object interactions. In addition, since our method incorporates MANO-based hand reconstruction, we compare our method with recent state-of-the-art hand mesh reconstruction methods in Table \ref{hand_recon}. Our method consistently achieves the best results across both procrustes-aligned and non-aligned metrics. Specifically, it achieves the lowest PA-V-PE of 4.9 mm, and the highest PA-J-AUC of 89.8\%, PA-V-AUC of 90.0\%, and PA-F@5 of 81.7\%, demonstrating precise articulation and high-fidelity mesh prediction. In the non-aligned setting, our method also outperforms previous approaches, reaching an F@15 of 94.4\%, significantly surpassing methods such as HandOccNet~\cite{park2022handoccnet} and H2ONet~\cite{akiva2021h2o}. These results confirm that our approach not only ensures joint-level accuracy, but also achieves detailed mesh reconstruction under complex occlusions and hand-object interactions.

\begin{table}[]
\renewcommand\arraystretch{1.25}
\centering
\caption{Ablation study on the masking ratio and the number of masking patches in our occlusion-aware MAE on the HO3Dv2 dataset \cite{hampali2020honnotate}. Here, $\mu$ denotes the masking ratio within the object region, whereas $\rho$ represents the total number of masking patches. "All” indicates that the masking covers the entire object bounding box.}
\begin{tabular}{cc|ccc|cc}
\toprule[1.5pt]
$\mu$&$\rho$ & MJE  & STMJE & PAMJE & OME  & ADD-S \\ \hline
25\%&12               & 22.2 &  21.0 & 10.1  &40.2  & 14.8  \\
50\%&12 (ours)               & 21.8 &  20.5 & 9.8  & 39.3 & 14.2  \\
75\%&12               & 24.3 &  23.9 & 10.5  &44.6 &  17.3 \\
100\%&12              & 29.3 &  31.2 & 12.6  & 50.6 & 20.3 \\ \hline
50\%&4               &22.6  &21.9   &10.3   & 42.0 & 16.2  \\
50\%&8               &22.3  &21.6   &10.1   & 40.8 & 15.7  \\
50\%&12 (ours)               & \textbf{21.8} & \textbf{20.5}  & \textbf{9.8}   & \textbf{39.3} & \textbf{14.2}  \\
50\%&16              &22.4  &20.9   & 10.2  & 41.1 & 14.9  \\ \hline
100\%&All              &31.1  &30.5   & 13.6  & 96.3 & 36.7  \\ 
\bottomrule[1.5pt]
\end{tabular}
\vspace{-1em}
\label{table:mask_ratio}
\end{table}

\par \textbf{Quantitative comparisons on HO3Dv2 Dataset:} A comparison with existing methods on the HO3Dv2 dataset is provided in Table~\ref{ho3d}. Our method achieves the lowest MJE of 21.8 mm, STMJE of 20.5 mm, demonstrating its superior capability in recovering fine-grained hand articulations even under severe occlusions. Compared to the previous state-of-the-art Qi\textit{ et al.} \cite{qi2024hoisdf}, our approach further improves both MJE and STMJE, while maintaining competitive performance in PAMJE. Notably, our method also surpasses previous works in object pose estimation, achieving an OME of 39.3 mm and the ADD-S of 14.2 mm. These results highlight the effectiveness of our framework in learning occlusion-aware representations that jointly enhance hand and object pose estimation. In addition, Table~\ref{tab3} further demonstrates the robustness of our method across diverse object categories in the HO3Dv2 dataset. The qualitative comparison between our method and HOISDF \cite{qi2024hoisdf} on the HO3Dv2 dataset is illustrated in Fig. \ref{fig5}. The results show that our framework exhibits robustness and effectively handles severe occlusions, whereas HOISDF struggles to maintain accurate pose estimation. Overall, our method remains stable and reliable under challenging occlusion conditions.

\subsection{Ablation Studies}
\label{sec:ablation}
To analyze the contributions of different components in our framework, we performed several ablation experiments on the HO3Dv2 dataset \cite{hampali2020honnotate}.

\par \textbf{Effect of Target-Focused Masking and Image Reconstruction: } To isolate the effect of our occlusion-aware masking strategy, we retain the encoder and decoder framework, but remove the masking mechanism. As shown in the first row of Table \ref{table:main_part}, this leads to a noticeable performance drop across all metrics. In particular, PAMJE increases from 9.8 mm to 10.5 mm, and OME increases from 39.3 mm to 43.8 mm. By focusing the masking strategy on object-centric regions and reconstructing the masked images, this approach enhances the model occlusion-awareness and its structural understanding of hand-object interactions.

\par \textbf{Effect of Multi-Scale Feature Fusion for SDF Regression: } To evaluate the effectiveness of our multi-scale image feature fusion, we ablate the hierarchical aggregation mechanism and instead directly regress the SDF using only the final-layer decoder features. As shown in the second row of Table~\ref{table:main_part}, this simplification weakens the model capacity to capture fine-grained local geometry and spatial variations. These results demonstrate that leveraging multi-scale features provides prediction.

\par \textbf{Effect of Implicit and Explicit Geometric Aggregation:}
We assess the contribution of the implicit–explicit geometric aggregation module by removing it and retaining only the implicit SDF-based representation. As shown in the third row of Table~\ref{table:main_part}, this leads to a slight performance degradation across all evaluation metrics. The results highlight the importance of fusing implicit and explicit geometric cues, as this fusion effectively captures both fine-grained surface details and the global spatial structure, which is especially advantageous for accurate and occlusion-aware hand-object pose estimation.

\par \textbf{Effect of Noise Perturbation on Mesh-Sampled Surface Points: }Removing the Gaussian perturbation applied to mesh-sampled surface points leads to a moderate drop in all metrics, indicating that this perturbation effectively bridges the distribution gap between training and voxel-sampled inference, allowing the network to learn smoother implicit fields.

\par \textbf{Effect of Backbone Substitution (DINOv2 → DINOv3) }: We evaluate the impact of the visual backbone by substituting DINOv2 with DINOv3 \cite{simeoni2025dinov3} under the same settings. As shown in Table \ref{table:main_part}, performance remains largely comparable, with a slight decline due to DINOv3 producing lower-resolution feature maps, which marginally reduces fine-grained spatial cues. This confirms the robustness and adaptability of our framework across different backbones.

\begin{table}[]
\renewcommand\arraystretch{1.25}
\centering
\caption{Ablation study on different masking types in our occlusion-aware MAE on the HO3Dv2 dataset \cite{hampali2020honnotate}.}
\begin{tabular}{c|ccc|cc}
\toprule[1.5pt]
\multicolumn{1}{c|}{Mask Type}   & MJE  & STMJE & PAMJE & OME  & ADD-S \\ \hline
Zero Masking     &22.5  &21.6   &10.2   & 42.1 & 16.3  \\
Mean Masking      &22.3 &21.3   &10.1   & 42.4 & 15.8  \\
Gaussian Noise (ours)  & \textbf{21.8}& \textbf{20.5}  & \textbf{9.8}   & \textbf{39.3} & \textbf{14.2}  \\ 
\bottomrule[1.5pt]
\end{tabular}
\vspace{-1em}
\label{table:masking_strategy}
\end{table}

\textbf{Effect on Loss Functions: }The ablation results of different loss components in our framework are summarized in Table~\ref{table:main_part2}. Removing the reconstruction loss $L_{\text{rec}}$, which belongs to our occlusion-aware MAE, leads to a moderate performance drop across all metrics, indicating that this term helps maintain feature consistency under occlusions. In contrast, removing the SDF supervision loss $L_{\text{SDF}}$ results in a drastic degradation across all metrics, confirming that the signed distance field provides an essential implicit geometric constraint for learning the hand–object interaction space. When excluding the MANO parameter regression loss $L_{\text{mano}}$, the model completely fails to estimate accurate hand poses, as reflected by extremely high errors in MJE, STMJE, and PAMJE. Similarly, removing the object pose loss $L_{\text{obj}}$ leads to an almost complete collapse in object localization (OME and ADD-S), demonstrating that this loss is indispensable for stabilizing the object pose branch. Overall, these results highlight that both $L_{\text{mano}}$ and $L_{\text{obj}}$ are fundamental for their respective pose estimation tasks, while $L_{\text{SDF}}$ and $L_{\text{rec}}$ provide crucial geometric and feature-level consistency to achieve optimal performance.


\par \textbf{Effect of Masking Ratio and Patch Number: }
We further study the influence of the masking ratio within the object region and the total number of masking patches. As shown in Table~\ref{table:mask_ratio}, the best performance is achieved when half of the object region is masked and twelve patches are sampled in total. A lower masking ratio provides insufficient supervision for occluded regions, while overly large masking ratios cause information loss and degrade reconstruction quality. Likewise, using either too few or too many patches slightly reduces performance, suggesting that a balanced masking strategy is crucial for effective occlusion reasoning and hand–object pose estimation. As shown in the last row of Table~\ref{table:mask_ratio}, when the entire object bounding box is masked, most object-related and some hand-related visual cues are occluded, leading to a severe drop in object pose accuracy and a slight decrease in hand pose performance.

\par \textbf{Effect of Masking Types: } We further conduct an ablation study on different masking strategies, including zero-masking, mean-masking, and Gaussian noise masking. As shown in Table~\ref{table:masking_strategy}, Gaussian noise masking achieves the best performance in all metrics. This suggests that during training, Gaussian noise introduces a more challenging reconstruction task, enabling the model to learn occluded region features more effectively and perform more robust reasoning. In contrast, zero-masking causes severe information loss in occluded areas, leading to suboptimal learning. Mean-masking retains some contextual information but lacks the variability needed for effective occlusion reasoning, making it less effective than Gaussian noise.

\begin{figure}
\centering
\includegraphics[width=1\linewidth]{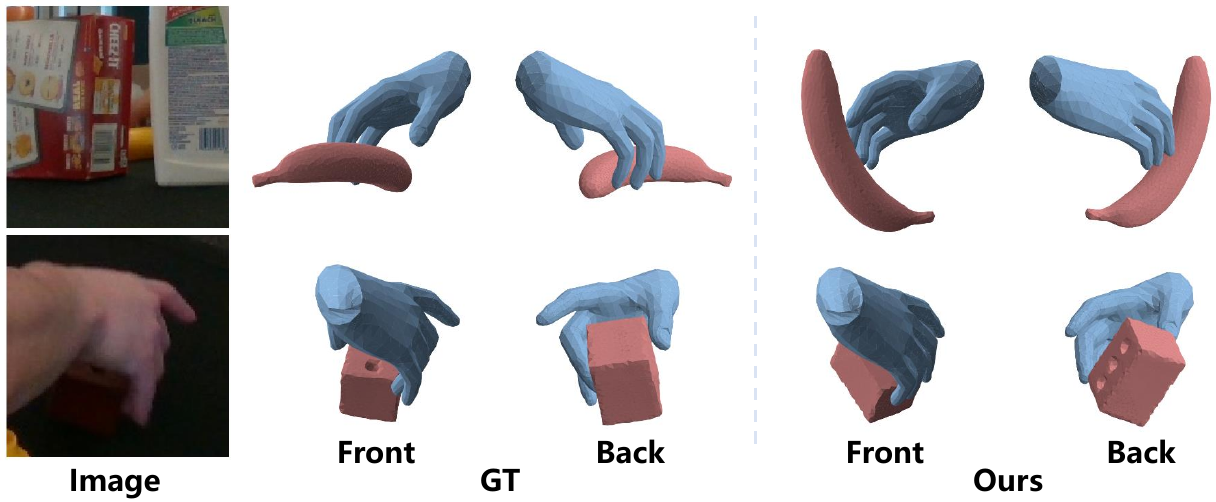}%
\caption{Failure cases of HOMAE. In scenarios with severe occlusions, the predicted hand-object poses exhibit inaccuracies.}
\label{fig6}
\vspace{-1em}
\end{figure}

\subsection{Computational Efficiency Analysis}
To assess the computational efficiency of our proposed framework, we evaluate the inference speed, parameter count, and memory consumption. The model achieves a real-time inference speed of 20 FPS, demonstrating its practicality for real-world applications. Despite incorporating the DINOv2 backbone and Transformer-based fusion modules, the total parameter count remains moderate at 206.9 MB, and the memory footprint is limited to approximately 1668 MB. These results indicate that our framework achieves a good balance between efficiency and accuracy, making it suitable for deployment in scenarios requiring real-time or near real-time hand–object pose estimation.

\subsection{Failure Cases and Limitations}

Although our method is robust in occlusion scenarios of hand-object interaction, it still has limitations under extreme occlusion conditions. Failure cases in extreme-occlusion interaction scenarios are shown in Fig. \ref{fig6}.
Specifically, when the object is completely occluded, the model faces challenges in accurately inferring its spatial relationships, often leading to imprecise pose estimations. Likewise, in cases of severe hand occlusion, incomplete feature perception may cause the model to misinterpret hand articulation. These limitations suggest that future research could further enhance the model robustness and spatial reasoning capability under extreme occlusion conditions through the integration of temporal cues or multi-view fusion strategies.

\section{Conclusion}
We introduced HOMAE, an occlusion-aware framework for 3D hand-object pose estimation from a single-view RGB image. By leveraging a target-focused masking strategy within masked autoencoders, our method enables context-aware feature learning and structural reasoning under occlusions. The integration of multi-scale SDF predictions with explicit point cloud representations further enhances geometric understanding, facilitating accurate and robust hand-object pose estimation. Our approach demonstrates strong generalization across complex interaction scenarios, as evidenced by state-of-the-art results on DexYCB and HO3Dv2. 

\textbf{Limitations and Future Work.} Although HOMAE demonstrates strong performance on benchmark datasets, its robustness decreases under extreme occlusion conditions. When both the hand and object are completely obscured, the model lacks sufficient contextual cues to support accurate pose estimation. Future work could investigate the temporal cues or multi-view fusion techniques to further enhance the robustness and accuracy in extreme occlusion scenes. In addition, we plan to investigate lightweight model architectures and deployment strategies for mobile or AR/VR devices, which would enable real-time inference and broaden practical applicability.

\bibliographystyle{IEEEtran}
\bibliography{reference}

\begin{IEEEbiography}[{\includegraphics[width=1in,height=1.25in,clip,keepaspectratio]{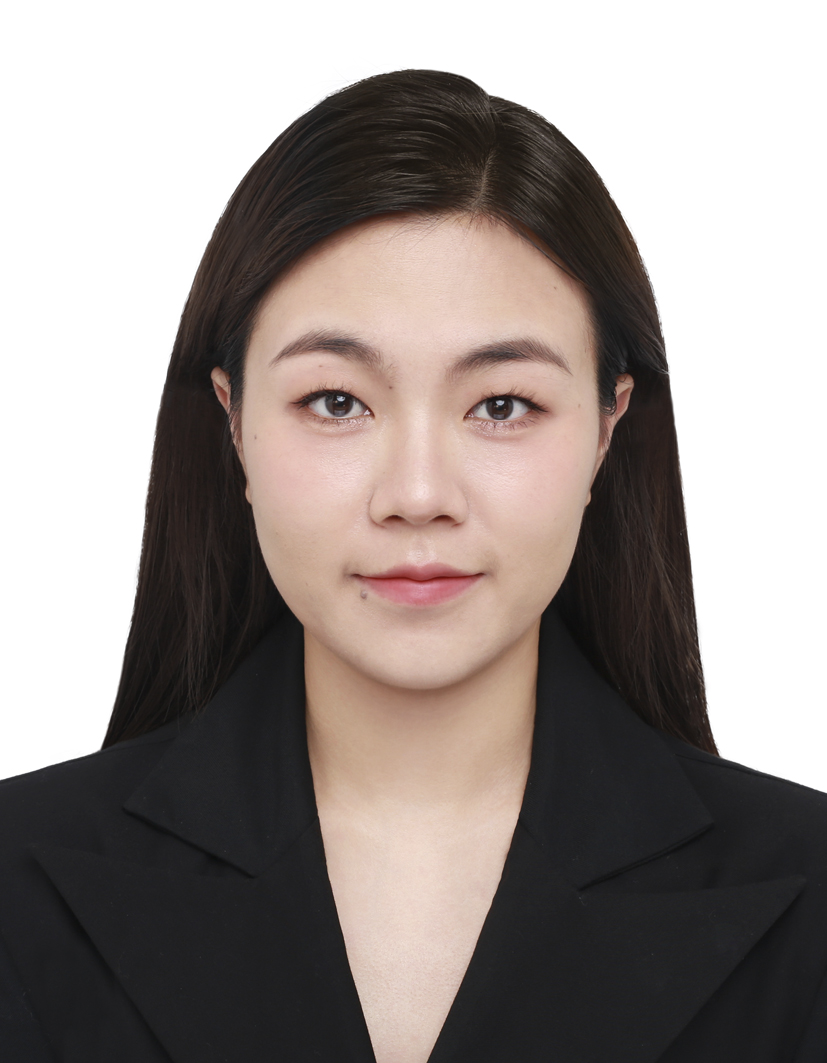}}]{Hui Yang}
received the B.S. degree from Dalian Maritime University, Dalian, China, in 2022. She is currently pursuing his Ph.D. degree at the National Engineering Research Center of Robot Visual Perception and Control Technology, School of Artificial Intelligence and Robotics, Hunan University, Changsha, China. Her research interests include hand-object pose estimation, point cloud analysis, and robotic manipulation.
\end{IEEEbiography}

\begin{IEEEbiography}[{\includegraphics[width=1in,height=1.25in,clip,keepaspectratio]{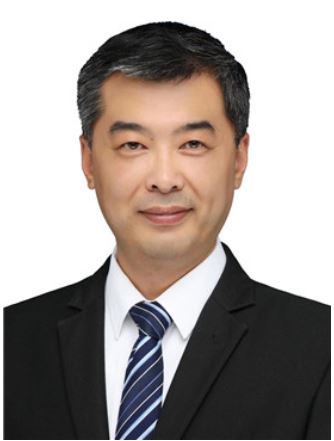}}]{Wei Sun} received the B.E. degree in industrial automation from Hunan University, Changsha, China, in 1997, the M.E., and Ph.D. degrees in control theory and control engineering from Hunan University, Changsha, China, in 1999, and 2003, respectively. He is currently a Full Professor at Hunan University and the Chief Scientist of the National Engineering Research Center of Robot Visual Perception and Control Technology. He received one Second-Grade National Technology Invention Award and two Second-Grade National Science and Technology Progress Awards of China. His research interests include robotics and artificial intelligence, with over 200 publications in these areas.
\end{IEEEbiography}

\begin{IEEEbiography}[{\includegraphics[width=1in,height=1.25in,clip,keepaspectratio]{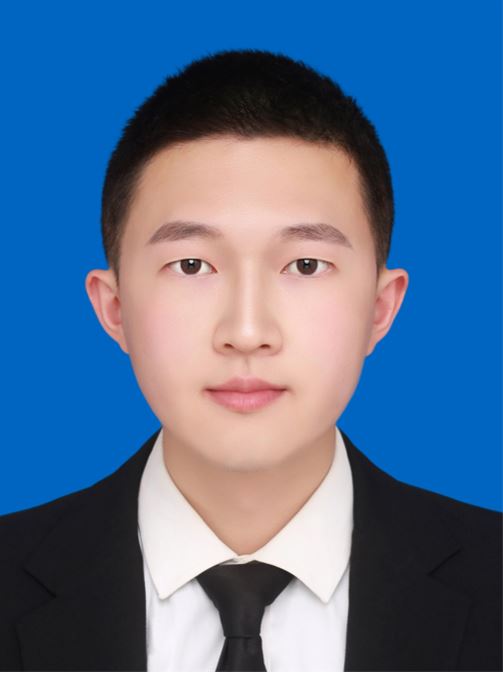}}]{Jian Liu} received his Ph.D. degree in 2025 at the National Engineering Research Center of Robot Visual Perception and Control Technology, Hunan University, Changsha, China. From 2023 to 2024, he was a Visiting PhD at the Department of Computer Science of the University of Western Australia. His current research interests include robotic manipulation, 3D machine vision, and object pose estimation. He served as a reviewer for more than 20 journals and conferences.
\end{IEEEbiography}

\begin{IEEEbiography}[{\includegraphics[width=1in,height=1.25in,clip,keepaspectratio]{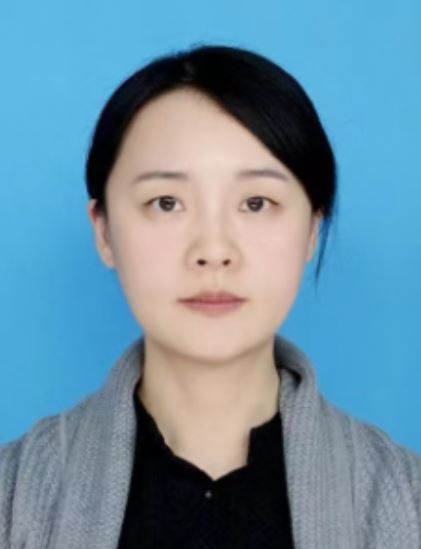}}]{Jin Zheng}
received the B.S., M.S., and Ph.D. degrees from the School of Architecture and Planning, Hunan University, Changsha, China, in 1998, 2001, and 2019, respectively. She is currently an Associate Professor with the School of Architecture and Art, Central South University, Changsha, China. Her current research interests include building intelligence and intelligent information processing.
\end{IEEEbiography}

\begin{IEEEbiography}[{\includegraphics[width=1in,height=1.25in,clip,keepaspectratio]{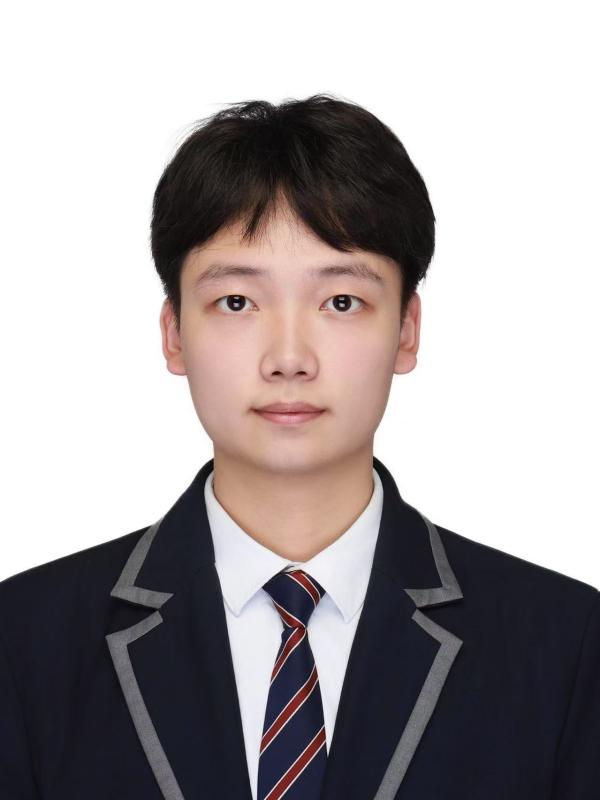}}]{Jian Xiao} 
received the B.S. degree in automation from the Hunan University, Changsha, China, in 2024. He is currently pursuing the Ph.D. degree with the National Engineering Research Center of Robot Visual Perception and Control Technology, College of Artificial Intelligence and Robotics, Hunan University, under the supervision of Prof. W. Sun. His current research interests include 3D computer vision, 6D object pose tracking, and imitation learning.
\end{IEEEbiography}

\begin{IEEEbiography}[{\includegraphics[width=1in,height=1.25in,clip,keepaspectratio]{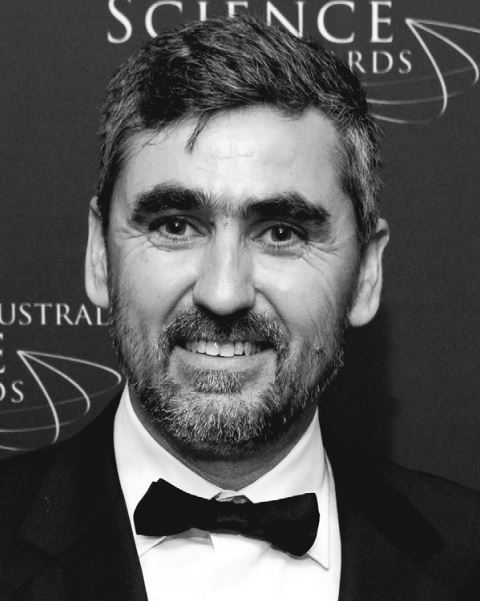}}]{Ajmal Mian} is a Professor of Computer Science at The University of Western Australia, Perth, WA, Australia. His research interests include computer vision, machine learning, and artificial intelligence. He has received the Vice-Chancellors Mid-Career Research Award and the IAPR Best Scientific Paper Award. He was an ARC Future Fellow, an IAPR Fellow, an ACM Distinguished Speaker, and the President of the Australian Pattern Recognition Society. He frequently serves as the AE/AC for several top-tier venues like IEEE TPAMI, CVPR, etc.
\end{IEEEbiography}

\end{document}